\documentclass[sigconf, nonacm, 10pt]{acmart}
\AtBeginDocument{%
  }
\usepackage{microtype}
\usepackage{graphicx}
\usepackage{subfigure}
\usepackage{booktabs} 
\usepackage{hyperref}
\RequirePackage{algorithm}
\RequirePackage{algorithmic}
\usepackage{amsmath}
\usepackage{multirow}
\usepackage{mathtools}
\usepackage{amsthm}
\usepackage[capitalize,noabbrev]{cleveref}
\usepackage{xspace}
\newcommand{\latinphrase}[1]{\textit{#1}}

\newcommand{\ie}{\latinphrase{i.e.,}\xspace}
\newcommand{\eg}{\latinphrase{e.g.,}\xspace}

\newcommand{\aka}{\latinphrase{a.k.a.}\xspace}

\DeclareMathOperator*{\argmax}{arg\,max}

\theoremstyle{plain}

\theoremstyle{definition}

\theoremstyle{remark}

\setcopyright{none}
\renewcommand\footnotetextcopyrightpermission[1]{} % removes footnote with conference information in first column
\newcommand\blfootnote[1]{%
  \begingroup
  \renewcommand\thefootnote{}\footnote{#1}%
  \addtocounter{footnote}{-1}%
  \endgroup
}
\begin{document}
\title{{Salted~Inference}:~Enhancing~Privacy~while~Maintaining Efficiency of Split~Inference~in Mobile~Computing}
\author{Mohammad Malekzadeh}
\affiliation{%
  \institution{Nokia Bell Labs}
  \city{Cambridge} 
  \country{UK} 
}
\email{{mohammad.malekzadeh@nokia-bell-labs.com}}

\author{Fahim Kawsar}
\affiliation{%
  \institution{Nokia Bell Labs}
  \city{Cambridge} 
  \country{UK} 
}
\email{fahim.kawsar@nokia-bell-labs.com}

\renewcommand{\shortauthors}{Mohammad Malekzadeh and Fahim Kawsar}

\begin{abstract}
In {\em split inference}, a deep neural network~(DNN) is partitioned to run the early part of the DNN at the edge and the later part of the DNN in the cloud. This meets two key requirements for on-device machine learning: {\em input privacy} and {\em computation efficiency}.  Still, an open question in split inference is {\em output privacy}, given that the outputs of the DNN are observable in the cloud. While encrypted computing can protect output privacy too, homomorphic encryption requires substantial computation and communication resources from both edge and cloud devices. 

In this paper\blfootnote{\normalsize To be appeared in the 25th International Workshop on Mobile Computing Systems and Applications (HotMobile'24), February 28--March 1, 2024, San Diego, CA, USA.}, we introduce {\em Salted DNNs}: a novel approach that enables clients at the edge, who run the early part of the DNN, to control the semantic interpretation of the DNN's outputs at inference time. Our proposed Salted DNNs maintain {classification \textit{accuracy} and {computation} \textit{efficiency}} very close to the standard DNN counterparts. Experimental evaluations conducted on both images and wearable sensor data demonstrate that Salted DNNs attain classification accuracy very close to standard DNNs, particularly when the {\em Salted Layer} is positioned within the early part to meet the requirements of split inference. Our approach is general and can be applied to various types of DNNs. {As a benchmark for future studies, we open-source our code at \textcolor{blue}{\mbox{\href{https://github.com/dr-bell/salted-dnns}{https://github.com/dr-bell/salted-dnns}.}}}
\end{abstract}




%%%%%%%%%%%%%%%%%
\maketitle

%%%%%%%%%%%%%%%%%
\vspace{.5cm}
\section{Introduction}\label{sec_intro}
Machine learning~(ML) has advanced {distributed and collaborative mobile} computing by enhancing the functionality of edge devices and introducing innovative applications {at the edge}. Running deep neural networks~(DNN) directly on edge devices, supported by robust cloud services,  facilitates real-time and low-latency ML inference for a diverse array of edge applications. However, for a variety of situations (\eg health monitoring, smart homes, or workplace environments), {\em clients} at the edge might not fully trust a {\em server} that provides an ML service to its clients. Thus,  clients may prefer to maintain the privacy of their personal data, {by keeping the data at the edge and preventing a server from accessing the data in its raw and unprocessed form}.

The idea of {\em split inference}~\cite{kang2017neurosurgeon} is to partition a DNN architecture such that some of the layers are {locally} run on clients' devices and the {rest of the layers} are centrally hosted on the server side, thus
splitting the computation between the client and the server. Split inference not only provides better {\em efficiency}~(compared to a client-only approach), but it also provides some level of {\em privacy}~(compared to a server-only approach). Notice that in a server-only approach, clients send their raw data, and the entire DNN is run on the server, while in a client-only approach, the entire DNN is run locally~\cite{kang2017neurosurgeon}. 

Imagine situations in which a client might want to {perform} multiple inferences on their private data. For example, a client might {use a smartphone to take} a picture of their skin {and share it} with {multiple} smartphone skin-analysis apps and combine the results of multiple {ML} services to make a more confident decision. 
In such situations, neither clients want to share their private data with all of these servers, nor are servers eager to completely offload their proprietary models to clients' devices. As a result, split inference addresses some important concerns on both sides and is considered to be one of the promising approaches. 

In the context of {this paper, an} encouraging feature of split inference is the fact that the DNN model employed in split inference could be trained using any arbitrary training procedure. The only requirement is that, after training, we need to split the model into two partitions: (1) the {\em early part} of DNN deployed at the client side, and (2) the {\em later part} of DNN deployed at the server side. Therefore, {\em split inference} is a concept more generic than {\em split learning}~\cite{gupta2018distributed}, which needs the model to be partitioned during training too~\cite{singh2019detailed, thapa2021advancements}. 

{Despite the benefits mentioned above}, the major weakness of split inference is that the {\em outputs} of the DNN are not private. Although the input data is kept private at the client side, the DNN’s output is observed by the server and is considered as public information by default. We can find several motivations for: \textit{why does one need to keep the outputs private too?} For example, when processing health or industrial data, protecting the privacy or business interests of the clients can be highly crucial. Especially, when detecting a sensitive disorder or disease or when classifying the nature of an anomaly or event in a workplace. {Other motivational} examples can be found in {everyday} recommendation or tracking applications, wherein the data of a client's behavior is processed {by ML models} to offer specific content or to take specific actions that might enhance the client's experience. Here, it is reasonable for clients to desire not only the privacy of their input behavioral data but also the confidentiality of the actions or contents recommended to them. {More} motivational examples can be found in private financial data, \eg estimating the credit score or mortgage allowance based on the client's personal data.

%%%%%%%%%%%%%%%%%%%%%%%%%%%%%%%%%%%
\begin{figure*}
    \centering
    \includegraphics[width=\textwidth]{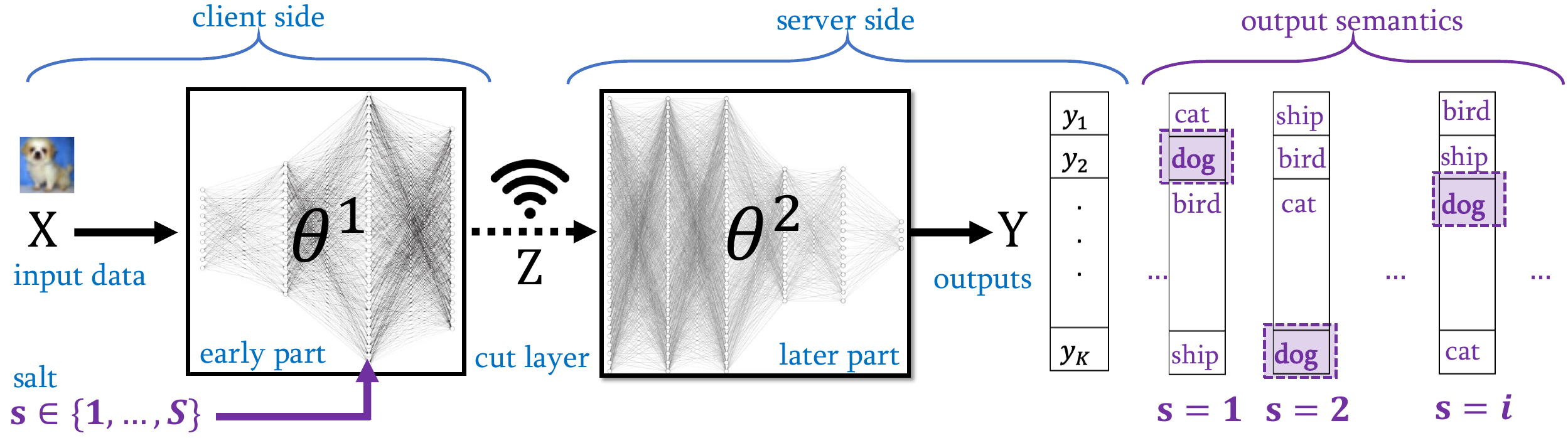}  
    \caption{The overview of our salted DNN for efficient and private split inference. The early part of DNN $\theta^1$ processes input data $X$ on the client's device. After processing up to the cut layer, the intermediate representations $Z=\theta^1(X)$ are transmitted to the server for further computation using the later part of the DNN. The DNN's outputs $Y=\theta^2(Z)$ are transmitted back to the client. The client controls the semantic arrangement of the outputs through the selected salt value $s$, making it the only party capable of decoding the meaning of the outputs.}
    \label{fig_big_pic}
\end{figure*} 
%%%%%%%%%%%%%%%%%%%%%%%%%%%%%%%%%%%

In this paper, we address {the following} open question: \textit{How can we achieve efficient split inference without revealing the true outputs to the server?} Specifically, we propose a solution that preserves the efficiency of standard split inference, while ensuring that the {semantics of} DNN’s outputs, calculated on the server side, are only interpretable by the client, and no other parties. 

\hfill

{\bf Contributions.} We introduce {\em Salted  DNNs}, {a novel approach to building DNNs where a randomly-generated {\em secret salt} is concatenated to a layer within the early part of the DNN to rearrange the {\em output's semantics}~(see Figure~\ref{fig_big_pic})}. We show that Salted DNNs can achieve classification accuracy close to standard DNNs, with only a marginal decrease of around 1\% to 3\% in accuracy. We show that positioning the {\em salted layer} within the early layers of the DNN results in remarkably high accuracy, aligning well with the requirements for output privacy in split inference. Remarkably, the salted layer introduced by our method is lightweight in terms of parameters and computations, making it highly efficient for split inference.

\section{Background and Objective}\label{sec_backg}
A motivation for split inference is that in the majority of real-world scenarios, some communications happen between clients and servers. Particularly, the majority of ML models running at the edge are trained, deployed, and maintained by ML service providers. 
A diagram of split inference is shown in Figure~\ref{fig_big_pic}, where at the client side, the early part of the DNN $\theta^1$ takes the input data and performs computations up to the {\em cut layer}. The outputs of the cut layer are transmitted to the server for further computation using the later part of the DNN $\theta^2$. Finally, the DNN's outputs are communicated back to the client. 

It is important to distinguish {\em split inference} from the concept of {\em split learning}~\cite{gupta2018distributed, singh2019detailed, thapa2021advancements}. In split learning, the outputs of the cut layer (which is the final layer of the early part) are shared during forward propagation, and only the gradients from the first layer of the later part are sent back to the client during backpropagation. In split inference, the outputs of the cut layer are sent to the server, but there is no need for backpropagation. A plethora of prior works have provided use-cases of split inference in edge and mobile computing~\cite{kang2017neurosurgeon, eshratifar2019bottlenet,  matsubara2019distilled, samragh2021unsupervised, banitalebi2021auto,  dong2022splitnets, matsubara2022split}. 

Split inference offers some level of privacy for input data: as a client does not share the original data $X$ (\eg a medical image), and only sends the intermediate features of the data $Z$ (\ie the representations of the data produced by the cut layer). One can even apply some careful regularization to $Z$ to prevent semi-honest servers (\aka honest-but-curious servers~\cite{chor1991zero, paverd2014modelling}) from reconstructing the original data~\cite{vepakomma2019reducing, song2019overlearning, malekzadeh2021honest, malekzadeh2022vicious}.  Nonetheless, split inference relies on a fundamental assumption: {\em the final outputs of the DNN are not considered private}. While the private input $X$ remains confidential on the client side, and there are methods to protect against the server's ability to reconstruct $X$ from $Z$, the requirement of ensuring {\em output privacy} remains critically unanswered. The output $Y$ is exposed to a  server and is typically treated as public information by default. 

{In \S\ref{sec_intro}, we presented motivating examples (such as healthcare, finance, and content recommendation) 
to illustrate the reasons why keeping the privacy of output $Y$ might be necessary. Taking a wider view, we can argue that input data $X$ is generally kept private due to the possibility of inferring sensitive information from it. Thus, if the input data $X$ is private, the same privacy concerns can be extended to any information $Y$ inferred from $X$. Particularly, DNNs are powerful in extracting information from data that is nontrivial and usually very difficult to extract using alternative methods.}

To protect the privacy of both inputs and outputs, a solution is {\em encrypted inference}, which protects both the private data of the client and the proprietary DNN of the server~\cite{gentry2013homomorphic, bourse2018fast}. However, running DNNs in an encrypted state imposes substantial computation and communication loads on clients and is impractical for edge devices where efficiency plays a critical role. Importantly, current techniques for encrypted inference are not compatible with the concept of split inference because they require heavy computations and multiple rounds of communication for every layer between client and server~\cite{hussain2021coinn}.

{Our objective is to enhance privacy protection in split inference by including the output privacy, while maintaining comparable levels of accuracy and computation efficiency. In split inference, the later part $\theta^2$ typically has a larger size compared to the early part $\theta^1$, given that servers have better computation resources. Our challenge, therefore, is to develop a solution that allows clients to achieve output privacy while enabling the server to perform the majority of computations, similar to standard split inference.}

%%%%%%%%%%%%%%%%%%%%%%%%%%%%%%%%%%%
\section{Our Methodology}

We propose {\em Salted DNNs}, a technique that enables us to {\em effectively} and {\em efficiently} control the semantics of the outputs produced by a DNN classifier. ``Effectively'' means to keep the accuracy of classifications very close to that of a standard  DNN. ``Efficiently'' means to maintain a similar level of computational demands as required by a standard  DNN. 

We introduce an algorithm for training Salted DNNs that enables clients to flexibly rearrange the DNN's output semantics during inference. Therefore, clients can effectively conceal the true inference outcome from the server. Our algorithm is generally applicable to a wide range of DNN architectures.

\subsection{Salted DNNs}\label{salted_dnn}

{\bf Inspiration}. {\em Salt} is a classic technique in cryptography to protect password hashes in a database against rainbow table attack~\cite{morris1979password}. In this technique, a randomly-generated string (known as {\em a salt}) is concatenated to the client's plain-text password, and then the compound string (\ie \mbox{\textit{salt + password}}) is hashed and stored in the database. The rationale behind {\em salt} is that, if two separate clients accidentally select the same password, their hashed passwords will still differ due to the use of different salts. 

{\bf Technique}. Inspired by the principle of a {\em salted password}, we introduce the concept of a {\em salted DNN}: we concatenate a randomly generated {\em secret salt} to a layer within the early part of the DNN to rearrange the {\em output's semantics}.  Our idea is to add a secondary input $s$ to the DNN that rearranges (or permutes) the output's order. The {\em secret salt} is a random variable drawn from a categorical distribution $s\in\{1,\dots, S\}$, such that without knowing the value of $s$, the server (or any other untrusted party) cannot make a reliable inference on the outputs of the DNN. That means, for any fixed input data $X$, the semantic order of the DNN outputs will depend on the chosen $s$.    In Figure~\ref{fig_big_pic}, we show an example of a salted DNN. After training the DNN with our proposed algorithm (presented in \S\ref{training}), the DNN is divided into two partitions: (1) the {\em early part} $\theta^1$ will be hosted at the client side and, (2) the {\em later part} $\theta^2$ at server side. In general, our method does not impose any restriction on the number or type of layers that can be used. 

{\bf Position}. Notice that, in split inference, the client should control the input $s$. Considering that $\theta^1$ can include $L$ layers, the input $s$ can be positioned before any layer, among layer $1$ to layer $L$. And in this setting, $\theta^2$ can include any number or type of layers. The position of input $s$ is a design decision that might affect the trade-off between the output secrecy (\ie privacy) and classification accuracy (\ie utility). Adding $s$ to early layers within DNNs aligns better with our objectives for private and efficient split inference. It also makes it harder for a server to infer the salt $s$ from $Z$ and, as a result, to decode the output semantics.

{\bf Explanatory Example.} Let input $X$ be an image and the classifier's outputs Y be the classifier's predictions with four possible categories \textit{$\{$1:cat, 2:dog, 3:bird, 4:ship$\}$}. Then, for a `dog' image, if $s=1$, the order of the classifier's outputs could be the default semantic order (\ie the second output shows the probability of being `dog'), and the softmax outputs could be something like
\begin{equation*}
\{\mathrm{y_1}, \mathbf{y_2}, \mathrm{y_3}, \mathrm{y_4}\} = \{0.05, \mathbf{0.6}, 0.2, 0.15\}.
\end{equation*}
These show that the probability of $X$ as a `dog' is $60\%$.  But, if $s=2$, then for the same input and the same DNN, the semantic order would change to \textit{$\{$1: ship, 2: bird, 3:cat, 4:dog$\}$}, and thus the outputs would look like 
\begin{equation*}
\{\mathrm{y_1}, \mathrm{y_2}, \mathrm{y_3}, \mathbf{y_4}\} = \{0.15, 0.2, 0.05, \mathbf{0.6}\}.
\end{equation*}
Similarly, the probability of `dog' is $60\%$ here, but now in a different order. Without knowing the value of $s$, and only by looking at the outputs of the DNN, one cannot figure out outputs semantics, and hence cannot understand the true class; which is considered as private information (\eg without knowing if $s$ is equal to 1 or 2, it is not clear which output corresponds to the probability of 'dog').

\subsection{Training and Inference Algorithms}\label{training}

To implement the functionality elaborated in \S\ref{salted_dnn}, we should train a DNN that is flexible in changing the output's semantic order, while maintaining high accuracy in correctly classifying the input data in that corresponding semantic order. With this goal, we introduce a technique for training DNN classifiers that rearranges the output order based on the chosen salt. Our proposed training algorithm is generic and works for DNN classifiers of any type that can have applications in split inference and beyond. We then explain how a client performs inference on such salted DNN after training. 

{\bf Training.} Algorithm~\ref{alg_training} shows our training procedure for Salted DNNs. Let $\mathbf{1}_y$ for  $y\leq K$ denote a {\em one-hot} vector of size $K$ where all the entries are zero except the entry at index $y$ that is equal to $1$. For example, if $K=5$, then $\mathbf{1}_3=[0,0,1,0,0]$. During training, for each input $X$ (of any type and dimensions) with one-hot label $\mathsf{Y}=\mathbf{1}_y$, we randomly select $s \in \{1,\dots, S\}$ and accordingly change the outputs' semantic to a setting specific to the chosen $s$.  We assume a one-to-one {\em mapping} function 
\begin{equation}\label{eq_map_func}
    \mathcal{M}:(s,\mathsf{Y})\rightarrow \mathsf{Y}^s
\end{equation}
that, for each $s$, it takes the default $\mathsf{Y}$ and maps it to a new label $\mathsf{Y}^s$. A simple example of such a mapping function is a mathematical operation called {\em modulo} (usually shown by symbol \%): when $K=S$, for each $\mathsf{Y}=\mathbf{1}_y$ and $s$, the new label is $\mathsf{Y}^s= \mathbf{1}_{(y+s)\%K}$. 

We train the DNN to learn the correct semantic order for each pair of input $(X,s)$ and thus provide the corresponding outputs  $Y^s=\theta(X,s)$ such that they are consistent with the ground-truth label $\mathsf{Y}^s$. The technique can be applied to any DNN architecture as we only need to concatenate an embedding layer for input $s$ to one of the layers in the early part of the DNN as illustrated in Figure~\ref{fig_big_pic}.

%%%
\begin{algorithm}[t]
    \caption{Training of Salted DNNs}
    \label{alg_training}
\begin{algorithmic}[1]
    \STATE {\bf Require:} $D=\{(X,\mathsf{Y})\}_{i=1}^{N}$: a dataset including pairs of input and ground-truth label; $S$: number of possible salts; $\mathcal{M}:(s,\mathsf{Y})\rightarrow \mathsf{Y}^s$: a mapping function for changing the outputs' order; $\theta$: a randomly initialized Salted DNN with the addition of salt input; $E$: number of epochs; $B$: batch size.
    \FOR{ $e$ from $1$ to $E$}
        \FOR{$i$ from $1$ to $N/B$} 
            \STATE Randomly select a batch of $B$ pairs of $(X,\mathsf{Y})$ from $D$, without replacement
            \STATE Randomly select a batch of $B$ salts $s$ from $\{1, \cdots, S\}$
            \STATE For each $s$ and $\mathsf{Y}$ in the batch, compute the new label $\mathsf{Y}^s$ using $\mathcal{M}:(s,\mathsf{Y})\rightarrow \mathsf{Y}^s$
            \STATE For each $(X,\mathsf{Y}^s,s)$, compute the loss $\mathcal{L}(Y^s=\theta(X,s),\mathsf{Y}^s)$
            \STATE Update $\theta$ via computed gradients on the loss $\mathcal{L}$
        \ENDFOR
    \ENDFOR
    \STATE {\bf return} $\theta$
\end{algorithmic}
\end{algorithm}
%%%
\begin{algorithm}[t]
    \caption{Split Inference on Salted DNNs}
    \label{alg_split_inference}
\begin{algorithmic}[1]
    \STATE {\bfseries Require:} $S$: number of possible salts; $\mathcal{M}^{-1}:(s,\mathsf{Y}^s)\rightarrow \mathsf{Y}$: the reverse of mapping function used for training; $\theta = (\theta^1(\cdot,\cdot), \theta^2(\cdot))$: a trained salted DNN split into two parts;
    \\{\bf $\quad\quad\quad\quad\quad\quad$--------- (A) at client side---------}
    \STATE  Take the input data $X$ 
    \STATE Select and save a salt $s$ from $\{1, \cdots, S\}$
    \STATE Compute the outputs of the cut layer $Z=\theta^1(X,s)$, 
    \STATE Send $Z$ to the server
    \\{\bf $\quad\quad\quad\quad\quad\quad$--------- (B) at server side---------}
    \STATE Compute the (salted) outputs $Y^s=\theta^2(Z)$
    \STATE Return $Y^s$ to the client
    \\{\bf $\quad\quad\quad\quad\quad\quad$--------- (C) at client side---------}
    \STATE Map $Y^s$ into the correct class by the chosen salt $s$ and the reverse of the mapping function $\mathsf{Y}=\mathcal{M}^{-1}(s,\mathsf{Y}^s = \mathbf{1}_{\argmax(Y^s)})$.
\end{algorithmic}
\end{algorithm}
%%%

{\bf Inference.} After training $\theta$ using our proposed training procedure in Algorithm~\ref{alg_training}, we partition $\theta$ into two parts. (1) The early part $\theta^1$ that includes the salt input and computes up to the cut layer: $Z=\theta^1(X,s)$. (2) The later part $\theta^2$ that takes $Z$ and computes the salted outputs: $Y^s=\theta^2(Z)$. During inference, the client exclusively controls the secret salt $s$, thereby the client is the only party who can interpret the correct order for the outputs. The procedure of split inference on a salted DNN is elaborated in Algorithm~\ref{alg_split_inference}.

%%%%%%%%%%%%%%%%%%%%%%%%%%%%%%%%%%%%
\subsection{{Threat Model}}

{We consider the conventional setup of split inference~\cite{kang2017neurosurgeon} with two {\em honest} parties: a client and a server; which means that both client and server adhere to the algorithms in~\S\ref{training} without deviation. Our contribution to split inference is the introduction of Salted DNNs to be used instead of standard DNNs. As a result, in addition to having input privacy against honest servers, clients are also given output privacy against honest servers.}
 
{Note that, after receiving the intermediate features $Z$, a server may attempt to reconstruct the private input $X$, or to decode the secret salt $s$ and, consequently, the semantic order of $Y$. Such a server is known as {\em semi-honest} or {\em honest-but-curious}~\cite{chor1991zero, paverd2014modelling}. While it is very difficult to accurately reconstruct $X$ from $Z$, studies have shown that some high-level attributes of $X$ can be inferred from $Z$~\cite{vepakomma2019reducing, song2019overlearning, malekzadeh2021honest, malekzadeh2022vicious}.}
 {Without applying systematic modifications to $Z$, such as quantization or noise addition~\cite{samragh2021unsupervised}, input privacy in split inference is not formally guaranteed against semi-honest servers. One method of protection against a semi-honest server is to run the DNN under homomorphic encryption~\cite{gentry2013homomorphic, bourse2018fast, hussain2021coinn}. However, as discussed in \S\ref{sec_backg}, this solution is entirely at odds with the computational efficiency promised by split inference.}
 
{Although the salt $s$ is a variable that is independent of $X$, it is an input to the DNN. Therefore, a semi-honest server can perform similar reconstruction attacks to decode $s$ from $Z$. The privacy risk posed by a semi-honest server is not unique to this paper, and is still an open area of research in split inference. We believe that any technique to enhance input privacy for $X$ can also be applied to secure $s$. We defer this to future work; discussed in \S\ref{sec_lim_fut}.}

%%%%%%%%%%%%%%%%%%%%%%%%%%%%%%%%%%%%
\begin{figure*}[t]
    \centering    
    \includegraphics[width=.7\textwidth]{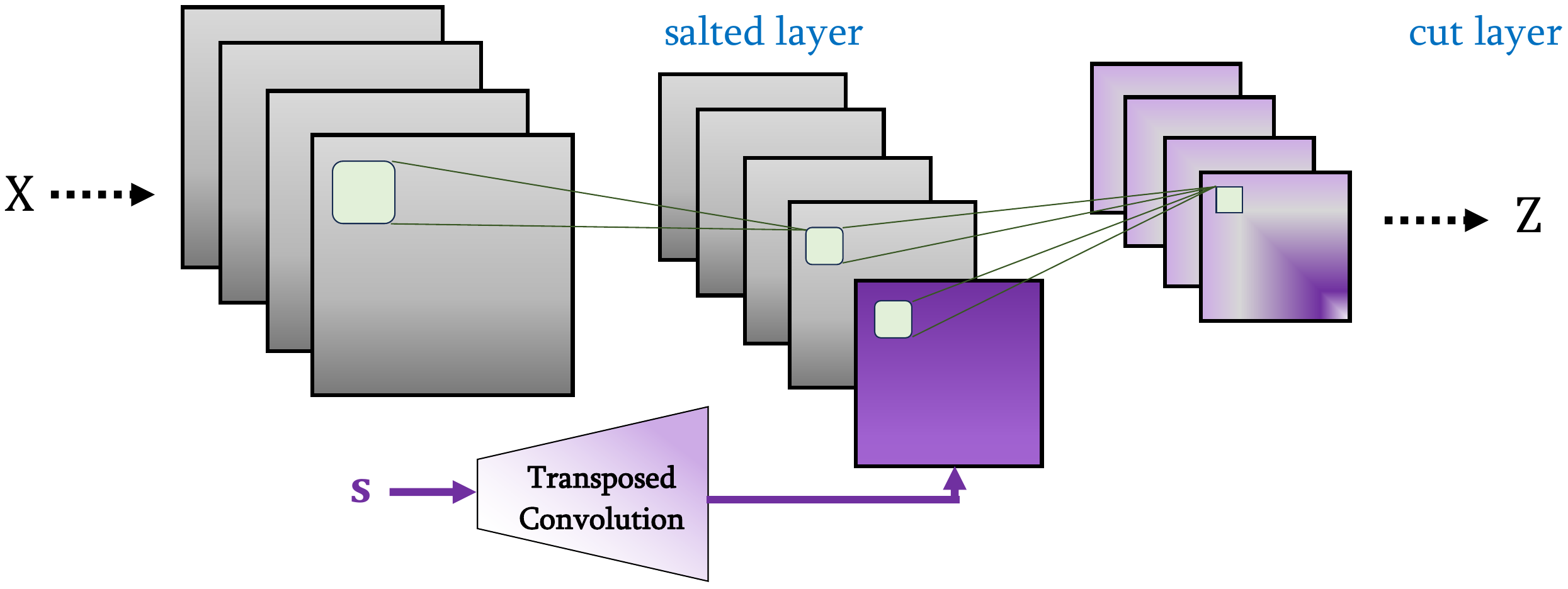} 
    \caption{A visual representation of a Salted Layer. A transposed convolutional layer expands the chosen salt $s$ into an output that matches the dimension of the salted layer.}
    \label{fig_salted_layer}
\end{figure*}

%%%%%%%%%%%%%%%%%%%%%%%%%%%%%%%%%%%
\section{Evaluation}\label{sec_eval}
\subsection{Experimental Set-up}
{\bf Datasets.} We evaluate Salted DNNs on two types of commonly-used datasets: image data from CIFAR10~\cite{krizhevsky2009learning} and time-series data of wearable sensors from PAMAP2~\cite{reiss2012introducing}; to represent typical private data capture by client devices. CIFAR10 has 10 classes and includes $50,000$ training images of dimensions $32\times 32\times 3$, in addition to $10K$ test images. PAMAP2 has 13 classes of physical activities, performed by 9 clients wearing 3 devices.  We use the training-test split provided in a benchmark library~\cite{PersonalizedFL}: $22,500$ training time-series, each with dimensions of $27\times 200$, in addition to $5600$ test time-series. Here, $27$ is the number of sensory readings from devices and $200$ is the length of each time series.

{\bf DNNs.} For CIFAR10, we use   \textbf{(i)}~LeNet~\cite{lecun1998gradient} (depth of 6 layers and total parameters of $126K$), and \textbf{(ii)}~WideResNet~\cite{Zagoruyko2016WRN} (depth of 28 layers organized into 3 residual blocks, each including 9 layers, and total parameters of $3.2M$). For PAMAP2 use \textbf{(iii)}~ConvNet~\cite{chang2020systematic} (depth of 9 layers and total parameters of $250K$). All DNNs consist of multiple convolutional layers followed by a few fully-connected layers. We use 500 training epochs with a batch size of 100, and Adam optimizer~\cite{kingma2014adam} with the default learning rate of $0.001$ and beta coefficients of $(0.9, 0.999)$. For mapping function $\mathcal{M}$, we use {\em modulo} mapping function $Y^s= \mathbf{1}_{(y+s)\%K}$, and thus we set $S=10$ for CIFAR10 and $S=13$ for PAMAP2, \ie $S$ is equal to number of classes $K$. For each experiment, we train the model for five iterations and report the mean and standard deviation of the classification accuracy on the test set. 

\subsection{Salted Layer Implementation}

As explained in Figure~\ref{fig_big_pic}, the salt input is concatenated to one of the layers within the DNN, which we refer to as the {\em salted layer}. To implement this, we must ensure that the dimensions of the salt input are aligned with the dimensions of the salted layer. We use {\em transposed convolution operation} which is an operation that increases the spatial dimensions of the input.  Therefore, the chosen salt $s$ is given as input to a {\em transposed convolution layer},  and the output of this transposed convolution layer is then concatenated to the input of the chosen salted layer within the DNN. Here, the design requirement is to make sure that the output dimensions of the transposed convolution layer match the input dimensions of the salted layer. As we show in Figure~\ref{fig_salted_layer}, and  Algorithm~\ref{alg_training}, the transposed convolution layer's parameters are trained to effectively expand the input $s$ into an output that aligns with the dimensions of the salted layer. More details of our implementations in PyTorch and reproducing our results are available in our repository at \textcolor{blue}{\mbox{\href{https://github.com/dr-bell/salted-dnns}{https://github.com/dr-bell/salted-dnns}}}.

%%%%%%%%%%%%%%%%%%%%%%%%%%%%%%%%
\subsection{{Evaluation Criteria}}

{To evaluate the performance of a Salted DNN, the first criterion is the classification accuracy on the test set. }
{Let $\mathbb{I}(\mathrm{C})$ denote the indicator function that outputs 1 if condition $\mathrm{C}$ holds, and 0 otherwise.  Given a test set $D=\{(X, s,\mathsf{Y})\}_{i=1}^{N}$, the {\em test accuracy} of a Salted DNN $\theta(\cdot,\cdot)$ is defined as:
\begin{equation}\label{eq_test_acc}
     \frac{1}{N}\sum_{i=1}^{N}  \mathbb{I}\big(\argmax(Y^s_i = \theta(X_i,s_i)) = \argmax(\mathsf{Y}_i^s) \big),
\end{equation} 
where $\mathsf{Y}_i^s= \mathcal{M}(s_i,\mathsf{Y}_i)$ as defined in \eqref{eq_map_func}.
}
{For each data point $X_i$, we {\em randomly} and {\em independently} sample a salt $s_i$. We then compute the corresponding salted outputs $Y^s_i = \theta(X_i,s_i)$. Next, we check if the predicted class from the salted outputs $Y^s_i$ is equal to the salted label $\mathsf{Y}_i^s$. \textit{A successful Salted DNN should have a test accuracy that is very close to the test accuracy of the standard DNN counterpart trained in the same setup.}}

{The second criterion is the ability to place the salted layer in various {\em positions}. This is important because in split inference for the same DNN architecture, the position of the cut layer can vary depending on the application or situation. If the cut layer is chosen to be layer $L$, the salted layer must be positioned before layer $L$. Therefore, we evaluate the test accuracy of a Salted DNN while concatenating the salted layer to different parts of the DNN. Note that when the salted layer is closer to the input, the chosen secret salt $s$ passes through more layers and non-linear operations before reaching the cut layer. This makes it more difficult for a semi-honest server to infer the salt $s$ from $Z$ and, as a result, to decode the output semantics. \textit{A successful Salted DNN should achieve a test accuracy very close to the standard DNN counterpart for various positions in the early part of the DNN.}}

{The third criterion is the {\em computation efficiency} of running a Salted DNN at inference time. Note that Salted DNNs do not add any extra communication overheads to split inference, because the data communicated between the client and the server remains the same as when using a standard DNN. However, there is a significant computational overhead introduced at training time, because Salted DNNs must learn to perform two tasks at the same time: data classification and output permutation. Since training occurs at the server, the cost of training a Salted DNN can be justified by the benefits it brings to clients. Therefore, we concentrate on computational overhead at inference time. \textit{A successful Salted DNN should have similar computational resource requirements to the standard DNN counterpart at inference time.}}

 {Taking these criteria into account, we evaluated the success of our design for Salted DNNs on various datasets and architectures. 
}

%%%%%%%%%%%%%%%%%%%%%%%%%%%%%%%%

\subsection{Results}
Table~\ref{tab_main_results} shows the main results of our experiments, and here we explain the key findings:

%%%%%%%%%%%%%%%%%%%%%%%%%%

{\bf (1) Test Accuracy.} For each of the three DNNs, we identified a reasonable position for the salted layer, enabling the Salted DNN to achieve a classification accuracy that closely matches that of the standard  DNN. For example, consider the results of the experiment in which the salted layer is the first layer of Block3 within the WideResNet (\ie the 18th layer of the overall 28 layers). The accuracy is $87.2\%$, only marginally lower by about 1\% compared to the standard counterpart, which achieved  $88.3\%$. 

{\bf (2) Position.} As previously mentioned, it is important for our purpose of private split inference to position the salted layer within the early part of the DNN; preferably closer to the input rather than the output. Again, for all three DNNs, we can achieve remarkably high accuracy when the salted layer is placed in the earlier layers, which aligns perfectly with our requirements. For example, the 3rd layer of the 9-layer Salted ConvNet or the 2nd layer of the 6-layer Salted LeNet show the highest accuracy.

{\bf (3) Efficiency.} In our experiments, we observed that, for all three DNNs, a Salted DNN requires approximately 2 to 3 times the number of epochs compared to the standard DNN to achieve convergence. Nonetheless, the key factor in split inference is the inference time efficiency, given that the training of a DNN occurs just once. After the deployment of a DNN, we need efficient inference on every input. As previously explained, a Salted DNN adds only {\em one} extra layer to the standard DNN, which is a very lightweight layer. For example, in Salted WideResNet, the additional transposed convolution layer introduces only $16K$ more parameters. This is remarkably small when compared with the $3.2M$ parameters of the standard WideResNet. Subsequently, the additional computations incurred by this extra layer are insignificant.

Our evaluations suggest that Salted DNNs can effectively satisfy the criteria for private and efficient split inference.

\begin{table}[]
\centering
\caption{Our main results as outlined in~\S\ref{sec_eval}. The Salted Layer is the position of the salt input, FC is a fully connected layer, and Conv is a convolutional layer. We bold the highest accuracy achievable by a Salted DNN when compared to a non-private standard DNN.}
\label{tab_main_results}
\resizebox{\columnwidth}{!}{%
\begin{tabular}{@{}lrcc@{}}
\toprule
\textbf{\textit{Dataset}} & \textbf{\textit{DNN Architecture}} & \textbf{\textit{Salt Position}} & \textbf{\textit{Accuracy (\%)}}\\ \midrule
\multirow{10}{*}{CIFAR10}
& LeNet
& {\em No Salt} & {\bf 68.1$\pm$0.7} \\\cline{3-4}
&\multirow{6}{*}{Salted LeNet} & Conv1 & 64.8$\pm$0.5 \\\cline{4-4}
&& Conv2 & {\bf 67.2$\pm$0.3} \\\cline{4-4}
&& Conv3 & 66.9$\pm$1.2 \\\cline{4-4}
&& Flatten &  64.1$\pm$0.4 \\\cline{4-4}
&& FC1 &  65.8$\pm$0.6 \\\cline{4-4}
&& FC2 &  63.5$\pm$1.3 \\\cline{2-4}
& WideResNet
& {\em No Salt} & {\bf 88.3$\pm$0.4}  \\\cline{3-4}
& \multirow{2}{*}{Salted WideResNet} & Block2 & 84.1$\pm$0.8 \\\cline{4-4}
&& Block3 & {\bf 87.2$\pm$0.6} \\\cline{1-4}
\multirow{4}{*}{PAMAP2}
& ConvNet
& {\em No Salt} & {\bf 92.1$\pm$0.3} \\\cline{3-4}
& \multirow{3}{*}{Salted ConvNet}
& Conv2 & 89.1$\pm$0.6  \\\cline{4-4}
&& Conv3  & {\bf 89.8$\pm$0.4} \\\cline{4-4}
&& Conv4  & 87.8$\pm$1.2 \\
\bottomrule
\end{tabular}%
}
\end{table}

%%%%%%%%%%%%%%%%%%%%%%%%%%%%%%%%%%%
\section{Limitations and Future work}\label{sec_lim_fut}

{Our proposed design for Salted DNNs shows intriguing properties and has promising applications in edge and mobile computing. There are many opportunities for future research on this topic.}

{\bf 1. Privacy Guarantee.}  As the chosen salt, $s$, is concatenated to a layer in the early part of DNN, extracting $s$ directly from the output of the cut layer, $Z$, is not a trivial task. The main reason is that multiple non-linear operations are occurring between the {\em salted layer} and the {\em cut layer}. Nevertheless, this paper does not offer a formal privacy guarantee regarding the possibilities of inferring $s$ from $Z$. In future work, one can explore potential attacks that a semi-honest server could perform to infer $s$. Moreover, one can explore the types of defenses that can be implemented in response to such attacks. {In particular, it would be valuable to better understand the connection between the position of the Salted Layer and the success of inference attacks on the secret salt.}

{\bf 2. Large Number of Classes.} In our results, we have noticed that for PAMAP2, which has 13 classes, the Salted DNN shows approximately a $3\%$ lower accuracy compared to CIFAR10, which has 10 classes and the reduction in accuracy is much less significant, around $1\%$. It is important to mention that we conducted similar experiments on CIFAR100~\cite{krizhevsky2009learning}, which contains 100 classes of images. However, we could not achieve significant accuracy on CIFAR100 with our chosen DNNs. We believe that for dealing with datasets containing a significantly greater number of classes, achieving high accuracy with Salted DNNs requires both more powerful DNNs and much larger training datasets. Notice that CIFAR100 and CIFAR10 both have $50K$ samples. In our future studies, one can go deeper into potential solutions for extending Salted DNNs to classify a greater number of classes.

{{\bf 3. Mapping Function and Training Procedure.} In our implementation, we considered only one type of one-to-one mapping function, $\mathcal{M}$, and one type of training loss function, which is cross-entropy for multi-class classification. A different mapping function might significantly influence the loss landscape and the training dynamics. Moreover, beyond classifiers, the structure and semantics of DNN outputs can change significantly, for instance for a regressor or an autoencoder. One can investigate this domain to discover mapping functions or loss functions that more effectively align with our proposed criteria for a successful Salted DNN.}

{\bf 4. Beyond Split Inference.} From a technical standpoint, our proposed Salted DNNs effectively convert a deterministic function (\ie standard DNNs) into a randomized function by introducing a random auxiliary input. This transformation enables the use of Monte Carlo sampling to estimate confidence levels, for instance, for uncertainty estimation in predictions made by Salted DNNs. On the other hand, it might be the case that a Salted DNN could be more robust to adversarial examples. Generating effective adversarial examples for Salted DNNs may be more challenging, as adversaries can manipulate the input data but have no control over the salt input at inference time. Overall, one can explore the prospective uses of Salted DNNs in other domains related to secure and efficient ML for edge and mobile computing.

\section{Conclusion}
A notable limitation of split inference is that the DNN outputs are not inherently private, potentially revealing sensitive information about clients. 
We introduced the novel idea of Salted DNNs, which obtains accuracy levels close to the standard while guaranteeing that DNN outputs remain exclusively interpretable by the client, effectively protecting output privacy in split inference. Our novel approach is an important step forward in improving the efficiency and privacy of ML in edge and mobile computing.
Our method applies to a wide range of applications, and it does possess certain constraints on the type of the model. For future studies, one can provide a formal privacy guarantee against semi-honest servers, expand the utility of Salted DNNs to datasets with a significantly large number of classes, search for more effective mapping and loss functions for training Salted DNNs, or explore the potential applications of Salted DNNs for other purposes in the machine learning community.

\begin{acks}
We thank Akhil Mathur for his valuable contribution and feedback on the initial concept of this paper. We thank Soumyajit Chatterjee for engaging in a productive discussion regarding this work.
\end{acks}
%%%%%%%%%%%%%%%%%%%%%%%%%%%%%%%%%%%
\bibliographystyle{ACM-Reference-Format}
\bibliography{refs}
\end{document}